\documentclass[letterpaper, 10 pt, conference]{ieeeconf}  
\pdfoutput=1

\IEEEoverridecommandlockouts                              

\usepackage{graphics} 
\usepackage{epsfig} 
\usepackage{amsmath} 
\usepackage{amssymb}  
\usepackage{comment}
\usepackage{graphicx}
\usepackage{url}
\usepackage{cite}
\usepackage{xcolor}
\usepackage{booktabs}
\usepackage{tabularx} 
\usepackage{multirow}
\usepackage{soul} 

\usepackage[utf8]{inputenc}
\usepackage[T1]{fontenc}

\newcolumntype{L}[1]{>{\raggedright\let\newline\\\arraybackslash\hspace{0pt}}p{#1}}

\title{\LARGE \bf
Sensor Visibility Estimation: Metrics and Methods for Systematic Performance Evaluation and Improvement
}

\author{
	Joachim Börger $^{1}$, Marc Patrick Zapf $^{2}$, Marat Kopytjuk $^{3}$, Xinrun Li $^{2}$, and Claudius Gläser $^{1}$ 
	\thanks{$^{1}$Robert Bosch GmbH, Corporate Research, 71272 Renningen, Germany 
		{\tt\small joachim.boerger@de.bosch.com}}%
	\thanks{$^{2}$Bosch (China) Investment Co. Ltd., Shanghai 200000, P.R. China  
		{\tt\small marcpatrick.zapf@cn.bosch.com}}%
	\thanks{$^{3}$Robert Bosch GmbH, Cross-Domain Computing Solutions, Germany}%
}

\begin{document}

\setlength{\abovedisplayskip}{1pt}

\maketitle
\thispagestyle{empty}
\pagestyle{empty}

\begin{abstract}

Sensor visibility is crucial for safety-critical applications in automotive, robotics, smart infrastructure and others: 
In addition to object detection and occupancy mapping, visibility describes where a sensor can potentially measure or is blind. 
This knowledge can enhance functional safety and perception algorithms or optimize sensor topologies.

Despite its significance, to the best of our knowledge, neither a common definition of visibility nor performance metrics exist yet. 
We close this gap and provide a definition of visibility, derived from a use case review. 
We introduce metrics and a framework to assess the performance of visibility estimators.

Our metrics are verified with labeled real-world and simulation data from infrastructure radars and cameras: 
The framework easily identifies false visible or false invisible estimations which are safety-critical.

Applying our metrics, we enhance the radar and camera visibility estimators by modeling the 3D elevation of sensor and objects. 
This refinement outperforms the conventional planar 2D approach in trustfulness and thus safety. 

\end{abstract}


\section{INTRODUCTION}
\label{sec:intro}

The \textit{visibility} of a sensor describes the area where it can or cannot collect data. 
Intelligent transport systems (ITS) and automated driving functions like trajectory planning, pedestrian protection and others need a dependable representation of the environment. 
The underlying perception algorithms rely on the visibility of the observed area for complete scene understanding: 
In particular, missing detections must reliably be traced back to true object absence (freespace) or a blind spot due to occlusion (potentially unsafe area).

\begin{figure}
	\centering
	\includegraphics[width=0.9\linewidth]{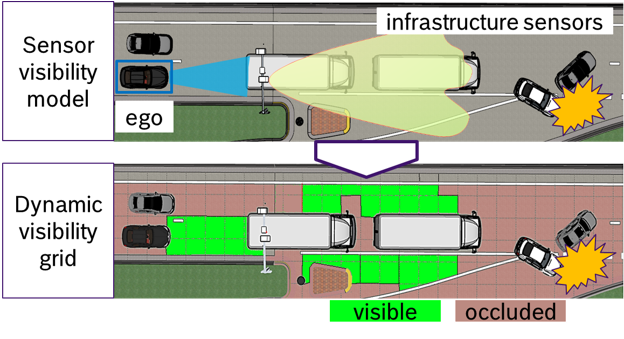}
	\caption{Motivation for visibility estimation. Top: An ego-vehicle receives perception input (e.g., obstacle detection) from onboard and infrastructure sensors. The FoVs (Field of View) are depicted in blue and yellow. Due to partial occlusion by trucks, the crashed vehicles at the highway on-ramp point are not visible to any sensor. This could wrongly be interpreted as \textit{freespace}. Bottom: The knowledge of visibility and occlusions serves as additional information for scene understanding and to warn of unsafe areas.}
	\label{fig:visibility_grid_concept}
\end{figure}

In Fig. \ref{fig:visibility_grid_concept}, a vehicle drives on a highway, with onboard object detection and long-range information from roadside sensors. The trucks obstruct all sensors, such that the accident at the on-ramp location cannot be detected. A \emph{visibility map} estimating visible and occluded areas can be used to mark the merge-in as unsafe area. To increase safety, the vehicle speed can be reduced, and the driver alerted.

Despite the importance -- any application using environment data should know  where data can be received -- there are way fewer works on visibility estimation than on classical challenges like object detection, tracking, segmentation etc. 

Visibility was formalized for sensor placement optimization \cite{10.1145/2906148}, terrestrial laser scanning \cite{Lecigne2020}, and autonomous agent navigation. However, these approaches and metrics are application-specific and hard to transfer to the automotive domain.

The main shortcomings of the state of the art are: Firstly, there is no common definition of visibility and its ground truth. Secondly, no universal metrics exist for a systematic evaluation or comparison of visibility estimators.

In this paper, we contribute:

\begin{enumerate}
	\item a formal definition of sensor visibility, derived from a review of relevant use cases
	\item a metric and evaluation strategy for the performance assessment of generic visibility estimators 
	\item an example evaluation of our visibility estimators on real-world data from infrastructure sensors
\end{enumerate}

\section{RELATED WORK}
\label{sec:related_works}


\subsection{Visibility estimation for cameras and sensor networks}
\label{subsec:rel_works_camera}

Visibility modeling has been studied to optimize sensor networks, predominantly for multi-camera setups.  \emph{Coverage} describes the area visible from a sensor system. This is either the ideal field of view (FoV) in absence of occlusions \cite{10.1145/1978802.1978811}, or \emph{partial coverage}, which is impacted by static or dynamic obstacles and often computed via line-of-sight evaluation. We call the latter \emph{visibility} for clear distinguishing.

Coverage is estimated with raytracing to optimize an infrastructure sensor setup for surveillance by Altahir et al. \cite{8054687} and for vehicle-to-everything (V2X) by Vijay et al. \cite{9564822}. 
They define coverage as the percentage of sensors that can see any given point on the surface. 
These works match our goal to define a sensor-agnostic visibility; however, the authors' objective is sensor placement at design-time, whereas our use cases require dynamic real-time data analysis.




\subsection{Visibility estimation for radar sensors}
\label{subsec:rel_works_radar}

Radar sensor modeling depends on numerous complex factors like object orientation \cite{Schipper2011} or radar cross-section \cite{Werber2015} which are often unknown in real-world applications. Current approaches span from phenomenological \cite{Bernsteiner2015} over deep learning-based \cite{Wheeler2017DeepStochasticRadarModels, weston2019probably} to physical  modeling \cite{Patole2017}. 

Inverse sensor modeling is required for occupancy map estimation; most applications like ours use efficient approximations of the aforementioned methods \cite{Werber2015, Slutsky2019, Li2018}.

Radar visibility estimation received little attention so far. 
Narksri et al. \cite{Narksri2021} estimate visibility for vehicle mounted sensors using a 3D map as external data source, focusing on hilly roads. 
Palffy et al. \cite{Palffy2019OcclusionAware} include occlusion information for pedestrian detection. However, the occluded area itself is detected by camera sensors, not radar.

\subsection{Evaluation of visibility estimation}

Por\k{e}bski and Kogut \cite{Porebski2021} propose an evaluation method for occupancy grids by assessing the correctness of the shapes of landmarks such as road signs. 

Schiegg et al. \cite{Schiegg2021} summarize metrics related to collective perception and communication. 
The \emph{environmental} or \emph{spatial awareness ratios} describe the ratios of objects or area perceived vs. missed by a communication node. 

Many visibility estimation approaches, ours included, are based on a preceding occupancy map. Weston et al. \cite{weston2019probably} evaluate occupancy map correctness using \emph{intersection over union}. Collins et al. \cite{Collins2007OccupancyGridMapping} propose a framework with three metrics: An image-based assessment of the map, a direct evaluation of obstacle detection based on it, and the usefulness of the map for path planning. 

The diversity of these approaches points out the importance of a good metric and serves as basis for our proposed visibility evaluation framework.

\section{USE CASES}
\label{sec:use_cases}

The following widely known use cases can benefit from visibility estimation, our metrics and evaluation framework:

\begin{figure}
	\centering
	\includegraphics[width=0.8\linewidth]{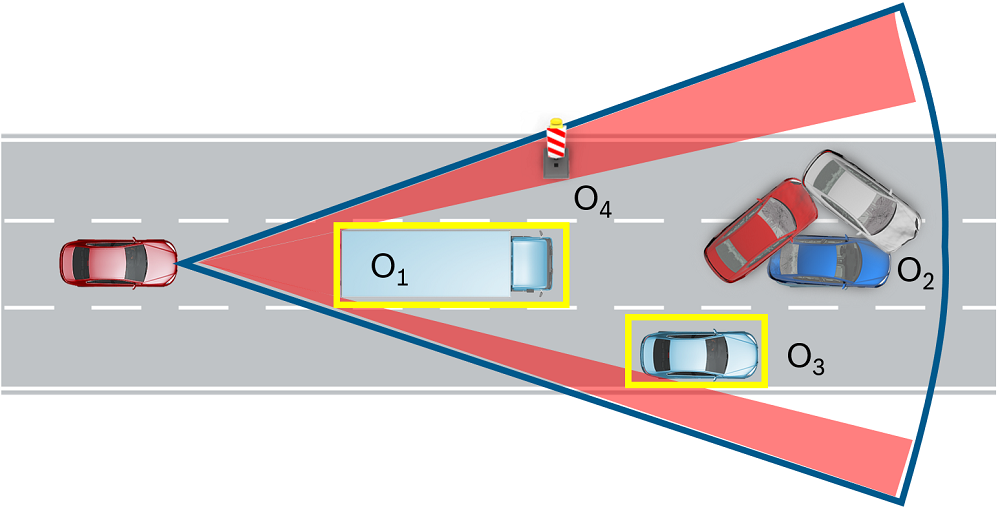}
	\caption{Example of FoV, visibility and detectability. The \emph{FoV} (blue) describes the maximum area where a sensor can detect objects given ideal conditions and free sight. \emph{Visibility} (red) represents all points with free line-of-sight to the sensor: $O_2$ is occluded by the truck $O_1$. \emph{Detectability (yellow)} considers additional objects properties (Table \ref{tableVisibilityFeatures}): $O_4$ is visible, but nondetectable due to its material or orientation. $O_3$ is occluded, but might still be detected by radar via beam reflection on the road-surface.}
	\label{fig:visibilitybirdview}
\end{figure}

\begin{itemize}
	\item Object tracking: Temporary occlusions can be compensated by predicting objects through occluded areas instead of dropping them. Spawning new objects can be sped up at the edge of the visible area, where they typically appear for the first time. Both aspects improve the calculation of existence probability \cite{aeberhard2017object}.

	\item Multi-sensor fusion: Imagine an object is detected by one, but not by another sensor. If this can be explained by occlusion, trust to the sensors is adapted according to their visibility to solve the contradiction.
	
	\item Sensor Monitoring: When monitoring false positive (FP) and false negative (FN) detections of one sensor compared to another, visibility can prevent wrong alerts.

	\item Safe trajectory planning: Freespace information can be used redundantly to object detection to verify that a planned trajectory is not blocked \cite{Michalke2020Where, Schiegg2019ObjectDP}. Naturally, occluded areas must be marked \emph{unknown} (Fig. \ref{fig:visibility_grid_concept}).
	
	\item Predicted visibility. Trajectory planning avoids driving into a blind area: Maneuvers which increase the predicted visibility can be prioritized, e.g., moving slightly towards the center line in a traffic jam situation.
	
	\item Sensor topology evaluation: A visibility estimator can predict the perception performance expected by a sensor setup \cite{Gabb2019InfrastructuresupportedPA}. The \emph{coverage rate} (Section \ref{subsec:statistical_evaluation}) measures system robustness or sensor redundancy.
	
\end{itemize}

The visibility information for these use cases can be summarized: 
\emph{Given location $(x, y)$ and assumed there is an unknown object $O$ -- can it be detected by the sensor or not?}

\section{DEFINITIONS}
\label{sec:definitions}
The above use cases need knowledge of \emph{Detectability}: the probability that an object $O$ is detected by the sensor. 
This not only requires a complex physical sensor model, but also information of the object's orientation, material etc., which usually is unknown in real-world applications. 

The term \emph{visibility} originates from Latin \emph{"visibilis" -- that may be seen}. 
We use this meaning to approximate detectability: It considers only the line-of-sight visibility of a spatial location by checking for occlusions. 
See Fig. \ref{fig:visibilitybirdview} and Table \ref{tableVisibilityFeatures} for a comprehensive summary.

\begin{table}
	\caption{FoV, visibility and detectability: Relation and aspects}
	\setlength\extrarowheight{2pt}
	\begin{tabularx}{\columnwidth}{p{4cm}|c|c|c}
		\toprule
		Aspects considered &  Detectability	& Visibility	& FoV  \\ 
		\midrule
		Constant physical sensor properties: Opening angle, max distances 	& x 	& x 	& x 	\\ 
		Location of the object 												& x  	& x 	& x  	\\ 
		Occlusion due to other objects 										& x 	& x 	& 	 	\\ 
		Orientation of the object  											& x 	&   	&   	\\ 
		Characteristics of the object (height, material, motion, ...) 	 	& x 	&   	&  		\\ 
		Weather conditions: Rain, fog, temperature, ... 					& x 	&   	& 	 	\\ 
		\bottomrule
	\end{tabularx}
	\label{tableVisibilityFeatures}
\end{table}

\subsection{Definitions}
\label{sec:DefinitionVisibilityDetectability}

Let $m = (x,y)$ represent a cell of a Cartesian grid, indexed by its upper left corner, and $O = (x, y, \phi, v_x, v_y, \phi')$ an object with position, orientation, velocity and yaw rate. We introduce the main definitions: 

\begin{itemize}
	\item \emph{Static Field of View} (FoV): Maximum area where a sensor can detect objects in ideal conditions, usually given by maximum detection distance, azimuth and elevation angle, w.r.t some reference object of a minimum size.
	
	\item \emph{Visibility of a cell}: $V_m = 1$ if some point $p$ within cell $m$ is visible from the sensor, i.e. the direct line of sight is not blocked. 

	
	\item \emph{Detection status} of an object: $D_O = 1 \Leftrightarrow $ at least one sensor measurement $z$ can be associated with $O$. Radar: The position of $z$ and Doppler velocity $v_D$ match $O$ within some tolerance. Lidar: Only the position matches. Camera: At least one pixel at the projected position of $O$ differs from the background.

\end{itemize}

\subsection{Definitions of basic terms}
\label{sec:DefinitionSimple}

Let $p = (p_x, p_y, p_z) \in \mathbb{R}^3$ be a point. For the definition of the features mentioned before, we also rely on:

\begin{itemize}
	\item Occupancy $Occ_m \in [0...1]$ describes the occupied ratio of a cell $m$. This is the inverse of freespace and required for the use case safe trajectory planning.


	\item Sensor \emph{measurement} $z$: A low level measurement, e.g., $z = (z_x, z_y, z_z, v_D)$. Usually, radars deliver 3D positions with Doppler speed $v_D$. Lidars or stereo cameras provide 3D positions with intensities, and mono cameras measure 2D pixel colors in the image plane.

	\item Visibility of an object: $V_O = 1$ if any cell $m$ overlapping with $O$ is visible, i.e. $\exists m: m \bigcap O \neq \emptyset, V_m = 1$ 
	
\end{itemize}



\section{METHODS}
\label{sec:methods}

The metrics defined and computed by our evaluation framework adhere to binary classifier evaluation such as \emph{true visible} or \emph{false invisible rate} and thus are generic and scalable. We apply our framework to various visibility estimators for radar and camera presented here. 

\begin{figure}
	\centering
	\includegraphics[width=0.8\linewidth]{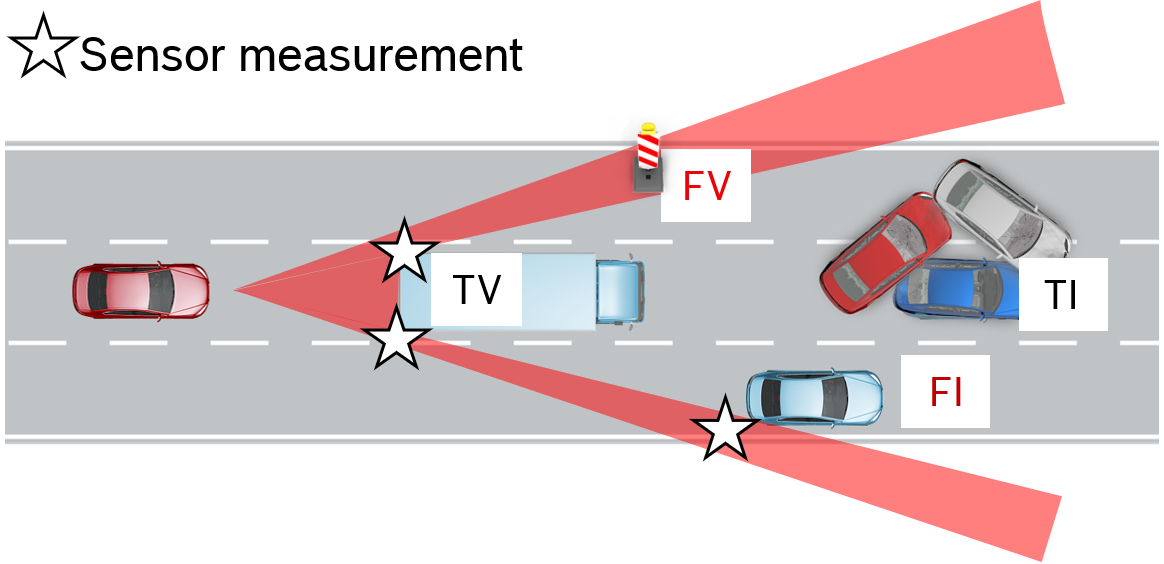}
	\caption{Performance metrics for visibility estimators: Analogue to binary classifier evaluation, the metrics measure the frequency of the four situations true visible (TV), false visible (FV), true invisible (TI), false invisible (FI).}
	\label{fig:visibility-object-kpis}
\end{figure}

\subsection{Metrics for the performance of estimated visibility }
\label{sec:proposed-kpis}

We consider algorithms that estimate the visibility $V_m$ for cells $m$. 
Recap that visibility approximates detectability, which is hard to compute and requires sensor and object knowledge. 
Still, the true \emph{detection status} of an object can be checked in measured data and will serve as \emph{gold standard}.

This idea implies that the evaluation becomes object-based rather than spatially-based, which fits perfectly to the use cases described before. 
Two criteria are relevant:

\begin{itemize}
	\item Robustness: For safe trajectory planning and object tracking, it is essential to detect invisible areas.

	\item Availability: At the same time, fearing too many blind spots reduces function availability; estimating too many occlusions therefore shall be avoided.
\end{itemize}

Table \ref{table:Metrics} illustrates the metrics to evaluate these criteria quantitatively. 
For each ground truth object $O$ and time step $t$, the comparison of the estimated object visibility $V_O$ to the true detection status $D_O$ yields one of the four situations TV, TI, FV, FI (Fig. \ref{fig:visibility-object-kpis}). 
With this terminology, we achieve analogy to the evaluation of binary classifier predictions. 
The frequencies are calculated as usual, e.g., false visible rate 

\begin{equation*}
	FVR = \frac{FV}{I} = \frac{FV}{FV + TI}
\end{equation*}

\begin{table}
	\caption{Metrics to assess the performance of visibility estimators}
	\setlength\extrarowheight{0pt}
	\begin{tabularx}{\columnwidth}{L{2.2cm} | L{2.2cm}  | L{2.8cm}}
		\toprule
															&  Prediction $V_O$: Visible	& Prediction $V_O$: Invisible  \\ 
		\midrule
		Ground truth $D_O$: Object detected by the sensor 		& True visible rate (TVR)					   		& False invisible rate (FIR): Decreased function availability 	\\ 
		& & \\
		Ground truth $D_O$: Object not detected 								& False visible rate (FVR): Unknown blind spots 	& True invisible rate (TIR): Correct detection of occluded areas	 	\\ 
		\bottomrule
	\end{tabularx}
	\label{table:Metrics}
\end{table}

\subsection{Conventional 2D radar visibility grid}
\label{sec:baseline-visibility}

A 2D radar visibility estimator is chosen as baseline. It computes the visibility $V_m^\text{Radar2D}$ for all cells $m$ in these steps:

\begin{itemize}
	\item Preprocessing: The raw data are filtered based on quality, elevation angle, RCS and radial velocity.
	
	\item Occupancy: A 2D Cartesian occupancy grid with resolution 1m~x~1m is processed with the filtered data: The inverse sensor model $p(Occ_m|z)$ estimates the occupancy of  $m$ given measurement $z$ by a scaled multivariate Gaussian probability density function (PDF) as in Elfes \cite{Elfes1989}. A \emph{decay function} accounts for dynamic objects as in Por\k{e}bski and Kogut \cite{Porebski2021}.	

	\item Visibility: $V_m$ is calculated based on $Occ_m$ by the line-of-sight approach similar to Adarve et al. \cite{Adarve2012}: Cells between the sensor origin and last occupied cell are marked visible, cells behind invisible. For directions without any occupied cells, the area up to the maximum FoV is assumed visible.
\end{itemize}

\subsection{Improved 3D radar visibility grid}
\label{sec:computation_radar_detectability}

With elevated sensor mounting positions, we now add height information in 3D to avoid false visible and invisible estimations (Section \ref{sec:results_radar_example_scenes}):

\begin{itemize}
	\item 3D spherical occupancy grid: Matching the nature of radar measurements, which provide a radial distance $r$, azimuth $\varphi$ and elevation angle $\alpha$, the internal Cartesian grid is replaced by a polar grid; for 3D modeling, it is enriched to a sphere by elevation information. Raytracing thus becomes very efficient, in particular for stationary sensors.

	\item To estimate a cells occupancy $Occ_m$ given a measurement in polar coordinates $z = (r, \varphi, \alpha)$, the dual inverse sensor model \cite{Slutsky2019} is applied. $p(Occ_m|z)$ is simplified to a scaled multivariate Gaussian PDF. 
	
	
\end{itemize}

\begin{figure}
	\centering
	\includegraphics[trim={0 0 0 20},clip,width=0.5\linewidth]{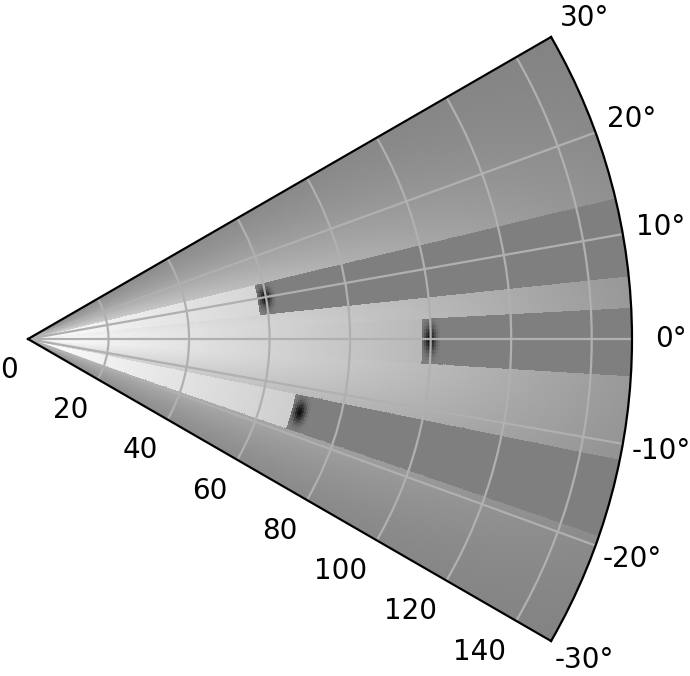}
	\caption{Spherical 3D radar visibility grid in polar coordinates (distance and azimuth angle), sliced to 2D as defined. Black shows occupied cells, white visible cells, and the intensity of gray depicts the grade of occlusion. }
	\label{fig:dual_ism_idea}
\end{figure}

\begin{figure}
	\centering
	\includegraphics[width=1.0\linewidth]{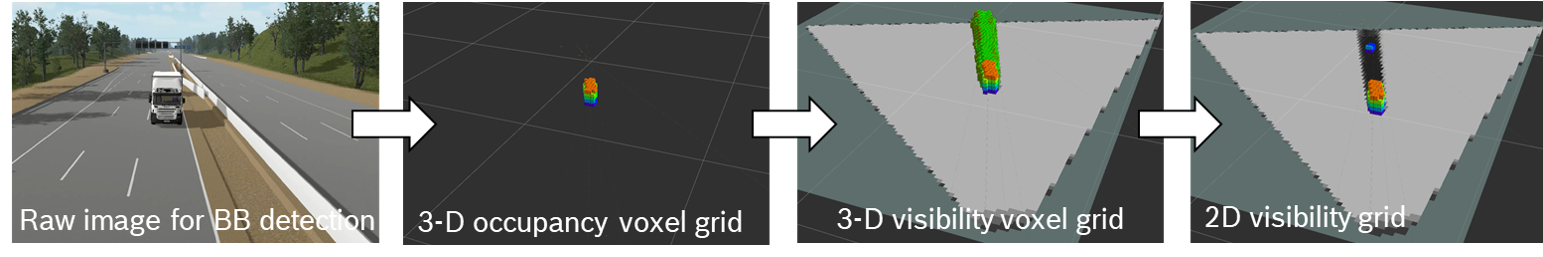}
	\caption{Creation of a camera visibility grid. 3D vehicle bounding boxes are estimated from raw camera images and voted into a 3D occupancy grid via homography projection. Using raytracing, a 3D visibility grid with visible (not shown) and occluded (green) voxels is created. For the final 2D output, the 3D visibility grid is squashed to 2D using z-averaging.}
	\label{fig:camera_grid}
\end{figure}

Since only the area of potential vehicle presence is relevant, the 3D grid is sliced at height $h=1$ meter above ground (Fig. \ref {fig:dual_ism_idea}). 
The final output $V_m^\text{Radar3D}$ on a 2D Cartesian grid, as defined in Section \ref{sec:DefinitionVisibilityDetectability}, is achieved by resampling.

The parameters (resolution, decay factor, positive and negative sensor models) were found using Bayesian optimization \cite{Madrigal2019} with training data recorded on a different day. 
	
Note that this visibility estimator is still very basic: 
The underlying occupancy grid is static, using a decay mechanism  for dynamics rather than an advanced technique like particle filters \cite{Danescu2011}. 
Still, the 3D enhancement achieves big improvements and is a good showcase for our visibility metrics.

\subsection{3D camera visibility grid}
\label{sec:computation_video_detectability}

A 3D visibility estimator based on raytracing is applied to infrastructure camera data (Fig. \ref{fig:camera_grid}). It follows our basic logic (Section \ref{sec:baseline-visibility}) in occupancy grid updating, decaying and visibility grid creation via line-of sight.

\begin{itemize}
	\item Vehicle bounding box estimation: Detection of 2D boxes via a CNN (YOLOv3) and subsequent 3D box estimation using vehicle size constraints and a homography projection from image space to world coordinates.
	
	\item Occupancy: A 3D Cartesian occupancy voxel grid is filled with the 3D data. Voxels covered by bounding boxes are updated as occupied.
	
	\item Visibility: 3D visibility voxel grids are computed using raytracing similar to radar (Section \ref{sec:computation_radar_detectability}).
\end{itemize}

We squash the 3D voxel grid to achieve the 2D output: For each cell $m$, the required $V_m^\text{Camera}$ is calculated as the average visibility value of all voxels above $m$ with height $h \in [0, 4]$ meters. This method slightly differs from radar due to better height measurements of the camera.

\subsection{Reference estimator with perfect occupancy grid}
\label{sec:visibility_with_perfect_occupancy}

To identify root causes of performance deficits of the visibility estimators, we recap the two main steps: 

\begin{enumerate}
\item Computation of the occupancy grid $Occ_m$
\item Estimation of visibility $V_m$
\end{enumerate}

For an isolated analysis, step 1) is replaced by a perfect occupancy grid created from ground truth object positions (Section \ref{sec:labeling-objects}). 
This can be interpreted as upper bound for the maximum achievable performance of a occupancy estimator. 
The evaluation then addresses only the visibility estimation (step 2) and reveals insights to the complexity and limitations of the underlying sensor model. 

The reference estimator is a generic concept. It was applied to the radar data in this investigation.



\section{EXPERIMENTS}
\label{sec:experiments}

We use labeled real world and simulation data to assess and improve the performance of the visibility estimators.


\subsection{Data description and sensor setup}
\label{sec:eval-setup-gt-data}

\begin{figure}
	\centering
	\includegraphics[width = 0.7 \columnwidth]{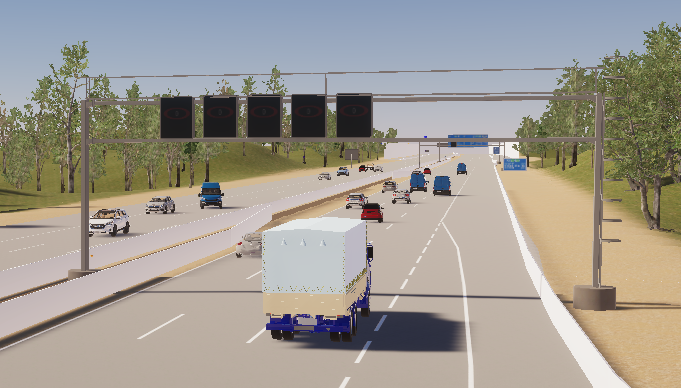}
	\caption{Real-world evaluation data is collected on a highway (depicted here as digital twin for privacy reasons). Radar and camera sensors are mounted on the gantry above the road, facing into the direction of travel.}
	\label{fig:eval-area-gmaps}
\end{figure}

The evaluation scene is a public motorway with three lanes in each driving direction (Fig. \ref{fig:eval-area-gmaps}). 
Sensor data was collected from one radar and one camera mounted on a gantry above the road in 6\,m height, facing into the direction of travel.

60 seconds of data with dense traffic and many occlusions from trucks were recorded at 18:45 evening time and labeled. 
The radar data contained 41 vehicle trajectories and is sufficiently representative to prove our evaluation method. 
For camera, due to privacy constraints, an equivalent 60-seconds timeseries from digital twin simulation was used.

\begin{figure}
	\hspace*{5.5mm}
	\includegraphics[width=0.777 \columnwidth]{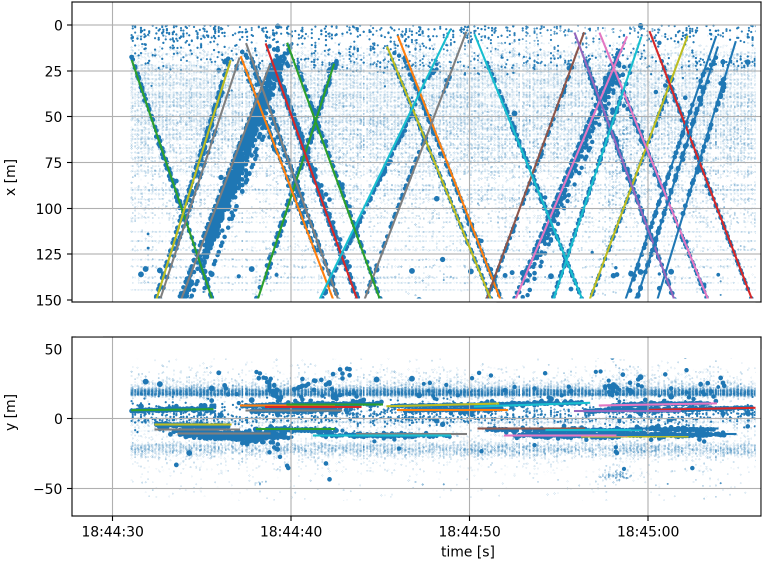}
	\caption{Labeling objects. Top: Radar reflections (dots) and labeled object trajectories (lines) with IDs are displayed with longitudinal position vs. time. Bottom: Lateral position vs. time. The raw radar data measure the object positions, so we just have to select those reflections corresponding to a vehicle and attribute an ID. To that aim, reflections with high Doppler velocities are plotted bold. Knowing the typical driving behavior of traffic participants in the surveilled area, we can easily select those points corresponding to one vehicle by drawing lines in a simple GUI.}
	\label{fig:trajectories_and_radar}
\end{figure}

\subsection{Labeling objects}
\label{sec:labeling-objects}

Thanks to our evaluation method, expensive highly precise labels can be skipped:
The ground truth detection status $D_O$ only requires knowledge of object presence close to the grid cells rather than exact object positions. 
This was achieved efficiently with the procedure described in Fig. \ref{fig:trajectories_and_radar}: Labeling of 1 minute data took roughly only 1 minute working time. 

Note that only moving objects can be labeled this way: The process relies on the visualization of the Doppler velocity; thus, stationary objects are hard to distinguish from clutter.

\section{RESULTS}
\label{sec:results_new}

For a demonstration of our metric, we present the performance assessed for the radar and camera visibility estimator. 

\subsection{Examples scenes}
\label{sec:results_radar_example_scenes}

\begin{figure}[tb]
	\hspace*{2mm} Radar True visible \hspace{13mm} True invisible  \vspace{1mm}
	\\
	\includegraphics[trim={0 0 50 15},clip,width=0.48 \linewidth]{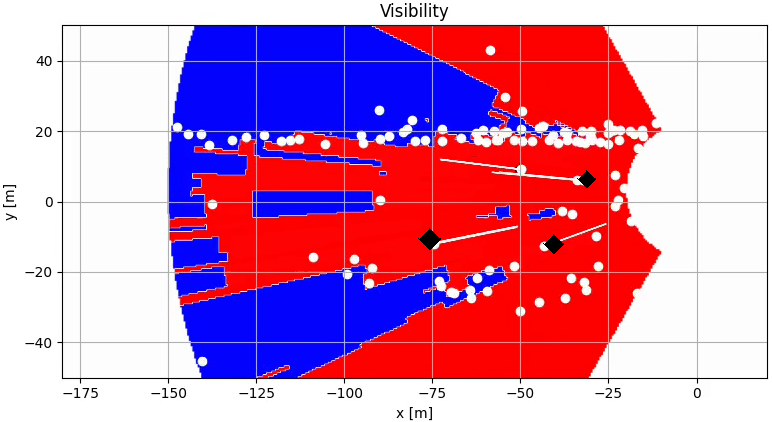}
	\includegraphics[trim={0 0 50 15},clip,width=0.48 \linewidth]{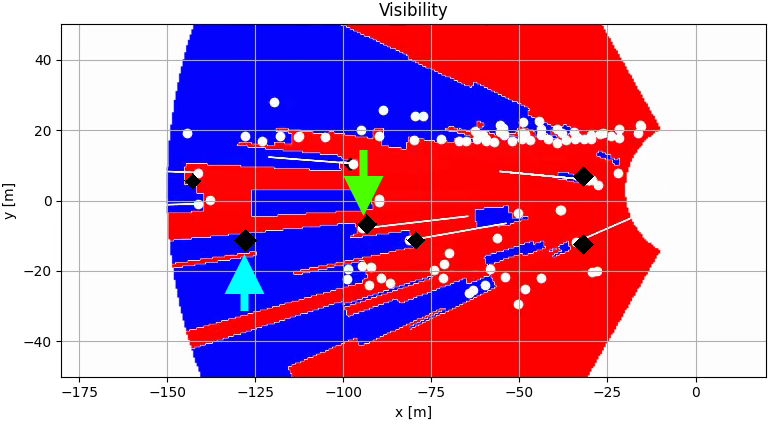}
	\\
	\hspace*{2mm} Radar False visible \hspace{13mm} False invisible  \vspace{1mm}
	\\
	\includegraphics[trim={0 0 50 15},clip,width=0.48  \linewidth]{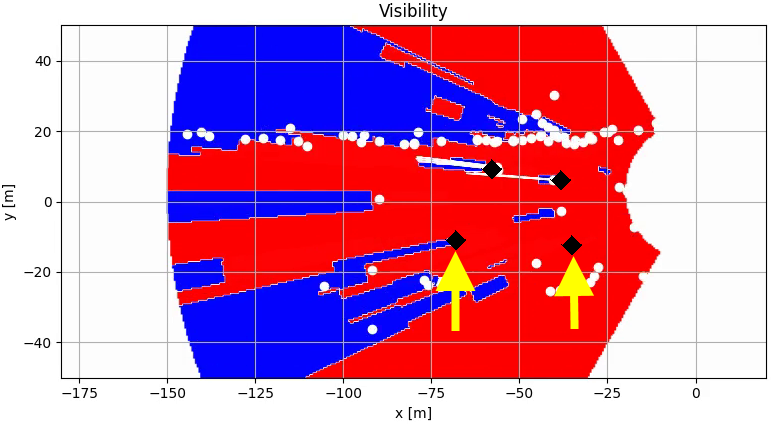}
	\includegraphics[trim={0 0 50 15},clip,width=0.48  \linewidth]{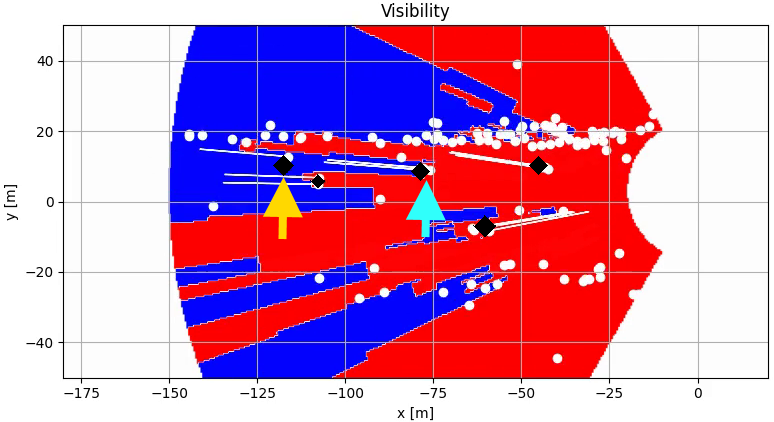}
	\caption{Estimated radar visibility (red), unknown (white) and invisible (blue). The radar is located at (0, 6) facing along the negative x-axis. Reflections $z$ are shown as white dots with Doppler velocity as lines. True visible: All objects (black diamonds) return a reflection and are correctly estimated visible. True invisible: The cyan object is occluded by the green one - it returns no reflection and is correctly invisible. False visible: The yellow objects return no reflection although in free line of sight to the sensor, which can have multiple root causes in the nondeterministic nature of radar (Section \ref{subsec:statistical_evaluation}). False invisible: The yellow object is detected (white dot), but estimated invisible because the cyan objects height is overestimated.}
	\label{fig:tp-example}
\end{figure}

\begin{figure}[tb]
	\hspace*{3mm} Camera False visible \hspace*{10mm} False invisible \vspace{1mm}
	\\
	\includegraphics[trim={0 0 0 0},clip,width=0.48 \linewidth]{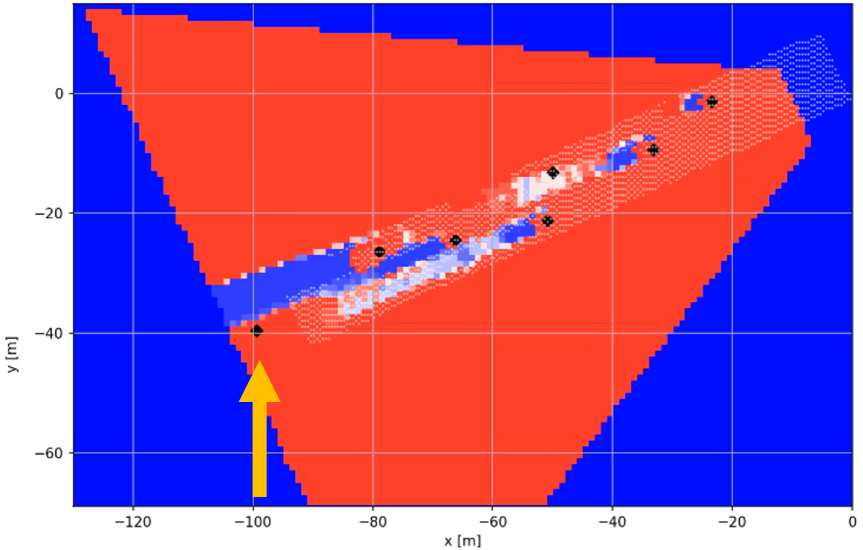}
	\includegraphics[trim={0 0 0 0},clip,width=0.48 \linewidth]{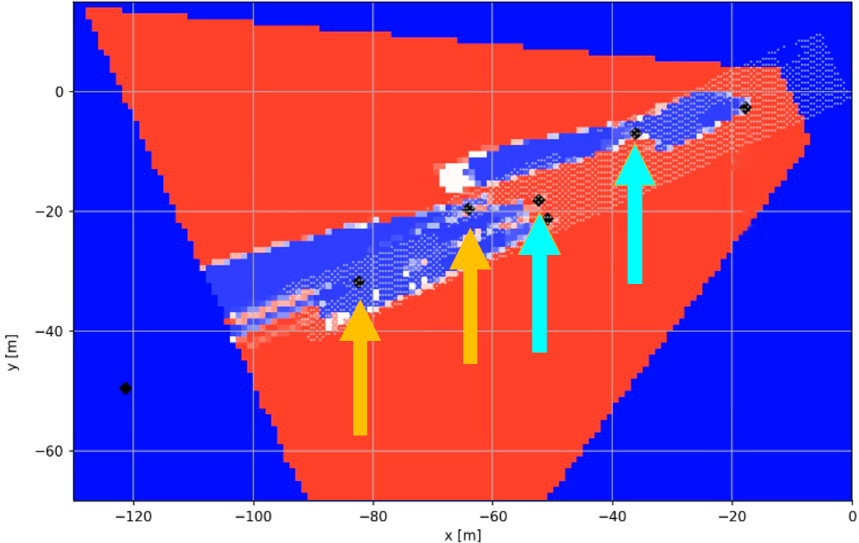}
	\caption{Estimated camera visibility (red), actual detectability area (faint white dotted area), invisible areas (blue) and ground truth objects (black diamonds). The camera is located at (0, 6) facing along the negative x-axis and slightly negative y-axis. 
False visible: The yellow object is outside the detection range (no faint white dot) but wrongly estimated visible. 
False invisible: The yellow objects are detected, yet wrongly estimated invisible because the cyan objects' heights are over-estimated, which results in too large invisible areas.}
	\label{fig:camera-example}
\end{figure}

With our evaluation framework, correct and wrong visibility estimations could easily be identified (Fig. \ref{fig:tp-example} and \ref{fig:camera-example}). 
For both radar and camera, the height over- or underestimation of vehicles often leads to FV or FI. 

Some radar FV stem from the absence of a reflection due to the nondeterministic nature of the sensor (Section \ref{subsec:statistical_evaluation}). Not depicted, also the underlying static occupancy grid sometimes reacts too slow, causing similar errors. 

At higher distance, the camera detector missed some vehicles. Thus, occluded space behind these vehicles could not be identified correctly, causing FV errors.

\subsection{Statistical evaluation of radar visibility}
\label{subsec:statistical_evaluation}

The resulting performance of the visibility estimators is summarized in Table \ref{tab:visibility-algo-results}. 

Radar achieves a total detection ratio for $D_O$ of 92.6\%, leaving 7.4\% undetected due to occlusion, averaged over all objects and time steps within the maximum static FoV. 
This number is interpreted as \emph{coverage rate} and can be regarded for optimization of the sensor position layout.

The improved 3D radar grid outperforms the 2D version by 36\% vs. 61\% FVR and thus achieves a higher trustfulness. 
The improvements stem from the height modeling of sensor and objects by considering the limited elevation FoV in near range, as well as the improved inverse sensor model.

However, the remaining FVR of 36\% still indicates too many undetected occlusions. 
Even the reference estimator with perfect occupancy information yields a FVR of 31\%. 
Thus, the major error source must be estimation of visibility rather than the preceding occupancy grid:
By construction, the underlying sensor models regard only line-of-sight visibility but no other aspects that influence the detectability (Table \ref{tableVisibilityFeatures}).

The error of 31\% thus quantifies the gap between the simplified line-of-sight visibility and true detectability (Section \ref{sec:definitions}).
It also ignores that the radar principle is nondeterministic by just evaluating binary visibility estimations versus binary detection results, which yields many FV without direct cause. We will comment on that in chapter \ref{sec:conclusions}.

The FIR improves from 23\% to 17\%: The 3D estimator models that elevated sensors can overlook obstacles after some distance, whereas the 2D estimator predicts an unlimited occlusion area behind the first obstacle. 

\subsection{Statistical evaluation of camera visibility}
\label{subsec:statistical_evaluation_camera}

The metrics of the camera visibility estimator were assessed on simulation data and are summarized in Table \ref{tab:visibility-algo-results}. 
The estimator achieves a good FVR of only 2\% due to overall dependable object bounding box detection. 
The FIR of 7\% mostly stems from over-estimation of bounding boxes, as well as voxel sizes in the 3D occupancy grid overestimating the actual size of the vehicles. 

Overall, the camera visibility estimator achieves a much better performance than radar. This comes from the deterministic nature of the camera to reliably detect objects. Additionally, the simulation data contain fewer trucks and occlusions, which explains the low FIR rate. This is subject to change in future investigations.



\begin{table}
	\centering
	\caption{Performance of the visibility estimators}
	\begin{tabularx}{\columnwidth}{p{5cm}|c|c|c}
		\toprule
		& TVR	&	FVR		&	FIR		\\ 
		\midrule
		Conventional 2D visibility grid 													& 77\% 	&	61\%	&	23\%	\\ 
		3D radar visibility grid 															& 83\% 	&	36\%	&	17\%	\\ 
		3D radar reference estimator with perfect occupancy grid										& 90\% 	&	31\%	&	10\%	\\ 
		&		&			&			\\
		3D camera visibility grid 															& 93\% 	&	2\%	&	7\%		\\ 
		\bottomrule
	\end{tabularx}
	\label{tab:visibility-algo-results}
\end{table}

%
%



\section{SUMMARY}
\label{sec:conclusions}



We contributed definitions of visibility applicable to various sensor modalities used in smart transportation systems. 
Our evaluation framework systematically assesses the performance and trustfulness of visibility estimators with \textit{run-time} data, which is important for safety-related applications.

The metrics were applied to radar and camera visibility estimators; the former could be augmented by 3D object and sensor height information and thus lower the FVR by 25\%.

We shed light on the importance of the underlying occupancy grids \cite{Thrun03} for the resulting visibility estimators: Their influence with an improvement potential of only 5-7\% is smaller than expected, which in turn emphasizes the importance of the sensor modelling part.

To avoid overestimating the FVR, the nondeterministic nature of sensors like radar should be considered: 
We propose to compare the estimated visibility probability against the empiric detection distribution rather than the binary detection event, e.g., via a Kolmogorow-Smirnow-Test.


\section*{ACKNOWLEDGMENT}
The authors would like to thank Thomas Nürnberg and Ronja König for thorough reviews and Peter Baumann, Nils Uhlemann and Adwait Kale for technical contributions.







\bibliographystyle{IEEEtran}
\bibliography{IEEEabrv,bibliography}

\end{document}